\documentclass[10pt,twocolumn,letterpaper]{article}

\usepackage{iccv}
\usepackage{times}
\usepackage{epsfig}
\usepackage{graphicx}
\usepackage{amsmath}
\usepackage{amssymb}
\usepackage{amsmath,bm}
\usepackage{multirow}
\usepackage{booktabs}
\usepackage[table]{xcolor}
\usepackage{algorithm}
\usepackage{algorithmic}
\usepackage{arydshln}
\usepackage{threeparttable}
\usepackage{colortbl}
\usepackage{enumitem}
\usepackage[numbers]{natbib}
\usepackage{array}
\usepackage[table]{xcolor}
\usepackage{colortbl}
\usepackage{caption}

\definecolor{mygray}{gray}{.9}
\definecolor{mygray1}{gray}{.7}
\newcommand{\tabincell}[2]{\begin{tabular}{@{}#1@{}}#2\end{tabular}}
\makeatletter
\newcommand{\thickhline}{%
    \noalign {\ifnum 0=`}\fi \hrule height 1pt
    \futurelet \reserved@a \@xhline
}
\usepackage[pagebackref=true,breaklinks=true,letterpaper=true,colorlinks,bookmarks=false]{hyperref}

 \iccvfinalcopy 

\def\iccvPaperID{2842} 
\def\httilde{\mbox{\tt\raisebox{-.5ex}{\symbol{126}}}}

\ificcvfinal\fi
\begin{document}

\title{Zero-Shot Video Object Segmentation via Attentive Graph Neural Networks}
\def\httilde{\mbox{\tt\raisebox{-.5ex}{\symbol{126}}}}
\author{ Wenguan Wang$^{1}\thanks{The first two authors contribute equally to this work.}$~,~~\hspace{1pt}Xiankai Lu$^{1*}$,~~Jianbing Shen$^{1}$\thanks{Corresponding author: \textit{Jianbing Shen}.}~,~~David Crandall$^{2}$,~~Ling Shao$^{1}$\hspace{1pt}   \\
	\small{$^1$} \small Inception Institute of Artificial Intelligence, UAE \hspace{0pt}
	\small{$^2$} \small Indiana University, USA \hspace{0pt} \\
	{\tt\small \{wenguanwang.ai, carrierlxk, shenjianbingcg\}@gmail.com}\\
 {\tt\small \url{https://github.com/carrierlxk/AGNN}}
	%
}

\maketitle
\thispagestyle{empty}

\begin{abstract}
   This work proposes a novel attentive graph neural network (AGNN) for zero-shot video object segmentation (ZVOS). The suggested AGNN recasts this task as a process of iterative information fusion over video graphs. Specifically, AGNN builds a fully connected graph to efficiently represent frames as nodes, and relations between arbitrary frame pairs as edges. The underlying pair-wise relations are described by a differentiable attention mechanism. Through parametric message passing, AGNN is able to efficiently capture and mine much richer and higher-order relations between video frames, thus enabling a more complete understanding of video content and more accurate foreground estimation. Experimental results on three video segmentation datasets show that AGNN sets a new state-of-the-art in each case. To further  demonstrate the generalizability of our framework, we extend AGNN to an additional task: image object co-segmentation (IOCS). We perform experiments on two famous IOCS datasets and observe again the superiority of our AGNN model. The extensive experiments verify that AGNN is able to learn the underlying semantic/appearance relationships among video frames or related images, and discover the common objects. 
\end{abstract}

\vspace{-6pt}
\section{Introduction}
	\vspace*{-4pt}	
Automatically identifying the primary objects in videos is an important problem that could benefit
a wide variety of  applications, by
reducing or eliminating manual effort needed to process and understand video.
However, 
discovering the most prominent and distinct
objects across video frames without having prior knowledge of what those
foreground objects are is a challenging task.
\begin{figure}[t]
  \centering
      \includegraphics[width=1 \linewidth]{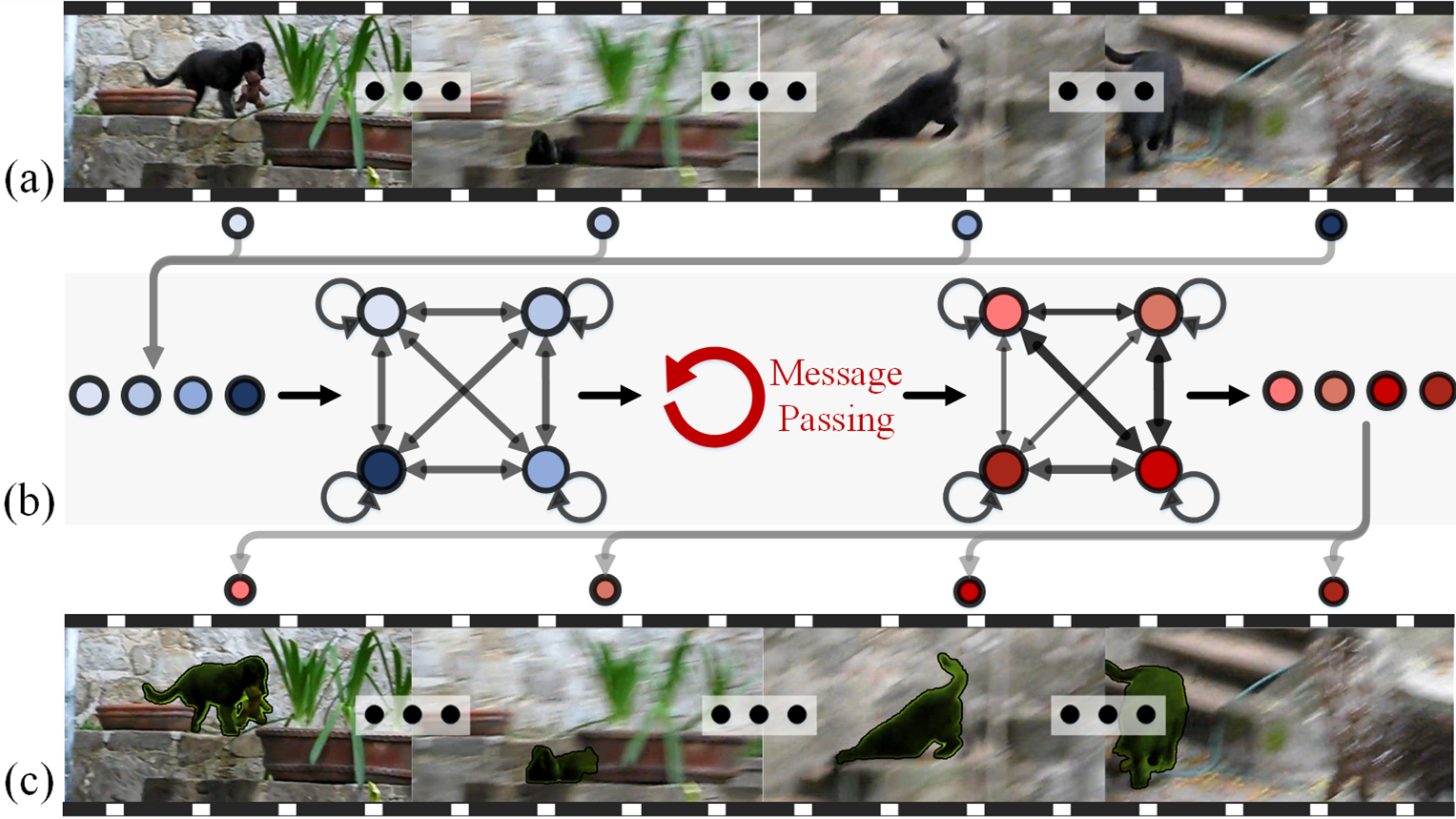}
\vspace{-18pt}
\caption{\textbf{\small Illustration of the proposed AGNN based ZVOS model.} \small (a) Input video sequence, typically with object occlusion and scale variation. (b) The suggested AGNN represents video frames as nodes (blue circles), and the relations between arbitrary frame pairs as edges (black arrows), captured by an attention mechanism. After several message passing iterations, higher-order relations can be mined and more optimal foreground estimations are obtained from a global view. (c) Final video object segmentation results. Best viewed in color. Zoom in for details. }
\label{fig:topright}
\vspace{-15pt}
\end{figure}
Traditional methods tend to tackle this issue by using handcrafted or learnable features in a \textit{local} or \textit{sequential} manner. For instance, handcrafted feature based methods use objectness~\cite{zhang2013}, motion boundary~\cite{DBLP:conf/iccv/PapazoglouF13}, and saliency~\cite{DBLP:conf/cvpr/WangSP15} cues over a few successive video frames, or explore trajectories~\cite{DBLP:conf/iccv/OchsB11}, \ie, link optical flow over multiple frames to capture long-term motion information. These are typically non-learning methods working in a purely \textit{unsupervised} manner. Recent deep learning based methods learn more powerful video object features from large-scale training data, yielding a \textit{zero-shot} solution~\cite{ventura2019rvos}  (still no annotation used for any testing frame). Many of these~\cite{cheng2017segflow,DBLP:conf/cvpr/TokmakovAS17,jain2017fusionseg,DBLP:conf/iccv/TokmakovAS17,Li_2018_ECCV1, Song_2018_ECCV} employ two-stream networks to combine local motion and appearance information, and apply recurrent neural networks to model the dynamics in a frame-by-frame manner.

Though these methods greatly promoted the development of this field and gained promising results, they generally suffer from two limitations. First, they focus primarily on the local pair-wise or sequential relations between successive frames, while ignoring the ubiquitous, high-order relationships among the frames (since frames from the same video are usually correlated). Second, since they do not fully leverage the rich relationships, they fail to completely capture the video content and hence may easily get inferior foreground estimates. From another perspective, as video objects usually suffer from underlying object occlusions, huge scale variations and appearance changes (Fig.~\ref{fig:topright} (a)), it is difficult to correctly infer the foreground when only considering successive or local pair-wise relations in videos.

To alleviate these issues, we need to explore an effective framework that can comprehensively model the high-order relationships among video frames into modern neural networks. In this work, an attentive graph neural network (AGNN) is proposed to addresses zero-shot video object segmentation (ZVOS), which recasts ZVOS as an end-to-end, message passing based graph information fusion procedure (Fig.~\ref{fig:topright}~(b)). Specifically, we construct a fully connected graph where video frames are represented as nodes and the pair-wise relations between two frames are described as the edge between their corresponding nodes. The correlation between two frames is efficiently captured by an attention mechanism, which avoids time-consuming optical flow estimation~\cite{cheng2017segflow,DBLP:conf/cvpr/TokmakovAS17,jain2017fusionseg,DBLP:conf/iccv/TokmakovAS17,Li_2018_ECCV1}.
By using recursive message passing to iteratively propagate information over the graph, \ie, each node receives the information from other nodes, AGNN can capture higher-order relationships among video frames and obtain more optimal results from a global view. In addition, as video object segmentation is a per-pixel prediction task, AGNN has a desirable, spatial information preserving property, which significantly distinguishes it from previous fully connected graph neural networks (GNNs).

AGNN operates on multiple frames, bringing the added advantage of natural training data augmentation,  as the combination candidates are numerous. In addition, since AGNN offers a powerful tool for representing
and mining much richer and higher-order relationships among video frames, it brings a more complete understanding of video content.
More significantly, due to its recursive property, AGNN is flexible enough to process variable numbers of nodes during inference, enabling it  to consider more input information and gain better performance (Fig.~\ref{fig:topright} (c)).

We extensively evaluate AGNN on three widely-used video object segmentation datasets, namely DAVIS$_{16}$ \cite{perazzi2016benchmark}, Youtube-Objects \cite{DBLP:conf/cvpr/PrestLCSF12} and DAVIS$_{17}$ \cite{pont20172017}, showing its superior performance over current state-of-the-art methods.

AGNN is a fully differential, end-to-end trainable framework that allows rich and high-order relations among frames (images) to be captured  and is highly applicable to spatial prediction problems. To further demonstrate its advantages and generalizability, we apply AGNN to an additional task:
image object co-segmentation (IOCS), which aims to 
extract the common objects from a group of semantically related images. It also gains promising results on two popular IOCS benchmarks, PASCAL VOC~\cite{pascal-voc-2012} and Internet~\cite{Rubinstein_2013_CVPR}, compared to existing IOCS methods.

Experiments on the ZVOS and additional IOCS tasks clearly demonstrate that AGNN is able to not only capture the relationships among correlated video frame images, but also mine the semantics among semantically related static images. Notably, this work can be viewed as a very early attempt to apply and extend GNNs for pixel-wise prediction tasks, which provides an effective video object segmentation solution and new insight into this task. 
	\vspace*{-5pt}
\section{Related Work}
	\vspace*{-4pt}	
\subsection{Graph Neural Networks}
	\vspace*{-4pt}	
\label{sec:gnn}
GNN was first proposed in~\cite{gori2005new} and further developed in~\cite{scarselli2009graph} to handle the underlying relationships among structured data. 
In~\cite{scarselli2009graph}, recurrent neural networks were used to model the state of each node, and the underlying correlation between nodes are learned via parameterized message passing over neighbors. Li \etal~\cite{li2015gated} further adapted GNN to sequential outputs. Gilmer \etal~\cite{gilmer2017neural} Later formulated the message passing module in GNNs as a learnable neural network. Recently, GNNs have been successfully applied in many fields, including molecular biology~\cite{gilmer2017neural}, computer vision~\cite{Qi_2018_ECCV,Wang_2018_ECCV,zheng2019reasoning}, machine learning~\cite{velickovic2017graph} and natural language processing~\cite{beck2018graph}. 
Another popular trend in GNNs is to generalize the convolutional architecture over arbitrary graph-structured data~\cite{duvenaud2015convolutional,niepert2016learning,kipf2016semi}, which is called graph convolution neural network (GCNN). 


The proposed AGNN falls into the former category; it is a message passing based GNN, where all the nodes, edges, and message passing functions are parameterized by neural networks. It shares the general
idea of mining relationships over graphs but has significant
differences. First, our AGNN is unique in its spatial information preserving nature, which is opposed to conventional fully connected GNNs and crucial for per-pixel prediction task. Second, to efficiently capture the relationship between two image frames, we introduce a  differentiable attention mechanism which addresses the correlated information and produces further discriminative edge features. Third, as far as we know, there is no prior attempt to explore GNNs in ZVOS.
	\vspace*{-3pt}
\subsection{Automatic Video Object Segmentation}
	\vspace*{-2pt}	
\label{sec:vos}
To automatically 
separate primary objects from the background, \textit{conventional} methods typically use handcrafted features (\eg, color, optical flow)~\cite{DBLP:conf/iccv/PapazoglouF13,DBLP:conf/bmvc/FaktorI14,tsai2016video,hu2018unsupervised} and certain heuristic assumptions related to the foreground (\ie, local motion differences~\cite{DBLP:conf/iccv/PapazoglouF13}, background priors~\cite{DBLP:conf/cvpr/WangSP15}). 
Some others explore more efficient object representations, such as dense point trajectories~\cite{DBLP:conf/iccv/OchsB11,DBLP:journals/pami/OchsMB14,wang2018semi} or object proposals~\cite{zhang2013,DBLP:conf/cvpr/KohK17,Koh_2018_ECCV,Lu_2018_ECCV}. Most of these methods work in a purely unsupervised manner without using any training data.

Recently, with the renaissance of deep learning, 
more research efforts have been devoted to tackling this in deep learning frameworks, leading to a zero-shot solution~\cite{fragkiadaki2015learning,jain2017fusionseg,DBLP:conf/iccv/TokmakovAS17,cheng2017segflow, Li_2018_CVPR,Li_2018_ECCV1,li2018flow,lu2019see}. For instance, a multi-layer perception based detector was designed in~\cite{fragkiadaki2015learning} to detect moving objectness. Li \etal~\cite{Li_2018_CVPR} integrated deep learning based instance embedding and motion saliency~\cite{Li_2018_CVPR} to boost performance. Some others turned to fully convolutional networks (FCNs)~\cite{cao2019triply,long2015fully,Ziqin2019RANet}. They introduced two-stream networks to fuse appearance and motion information~\cite{li2018flow,jain2017fusionseg,cheng2017segflow}, or explored more efficient feature extraction models and LSTM variants~\cite{Song_2018_ECCV}, to better locate the foreground objects.

The differences from previous methods are multifold: our AGNN \textbf{1)} provides a unified, end-to-end trainable, graph model based ZVOS solution; \textbf{2)} efficiently mines diverse and high-order relations within videos, through iteratively propagating and fusing messages over the graph; and \textbf{3)} utilizes a differentiable attention mechanism to capture the correlated information between frame pairs.
\vspace*{-3pt}
\subsection{Image Object Co-Segmentation}
	\vspace*{-2pt}	
\label{sec:coseg}
IOCS~\cite{rother2006cosegmentation,Mukherjee2009,Hochbaum2009} aims to jointly segment common objects belonging to the same semantic class in a given set of related images. Early methods usually formulate IOCS as an energy function defined over the whole or a part of the image set and consider intra- and inter-image cues~\cite{Vicente2010CosegmentationRM,Kim2011,Rubio2012,vicente2011object}.
To capture the relationships between images, some methods applied scene matching techniques
~\cite{Rubinstein_2013_CVPR}, global appearance models~\cite{wang2016higher}, discriminative clustering methodologies~\cite{joulin2010discriminative}, manifold ranking~\cite{quan2016object}  or saliency heuristics~\cite{han2018,tao2017image}.
There are only a very few deep IOCS models~\cite{chen2018semantic,DBLP:journals/corr/abs-1804-06423}, mainly due to the lack of a proper, end-to-end modeling strategy for this problem.  
\cite{chen2018semantic,DBLP:journals/corr/abs-1804-06423} tackled IOCS through a pair-wise comparison protocol and employed a Siamese network to capture the similarity between two related images.
Our AGNN based ICOS solution is significantly different from~\cite{chen2018semantic,DBLP:journals/corr/abs-1804-06423}. First, \cite{chen2018semantic,DBLP:journals/corr/abs-1804-06423} consider IOCS as a pair-wise image matching problem, while we formulate IOCS as an information propagation and fusion process among multiple images. That means our model can capture richer relations from a global view. Second, the Siamese network based systems only handle pair-wise relations, while our message passing based iterative inference can learn higher-order relations among multiple images. Third, our method is based on the graph model, yielding a more general and elegant framework for modeling IOCS.

\begin{figure*}[t]
  \centering
      \includegraphics[width=1 \linewidth]{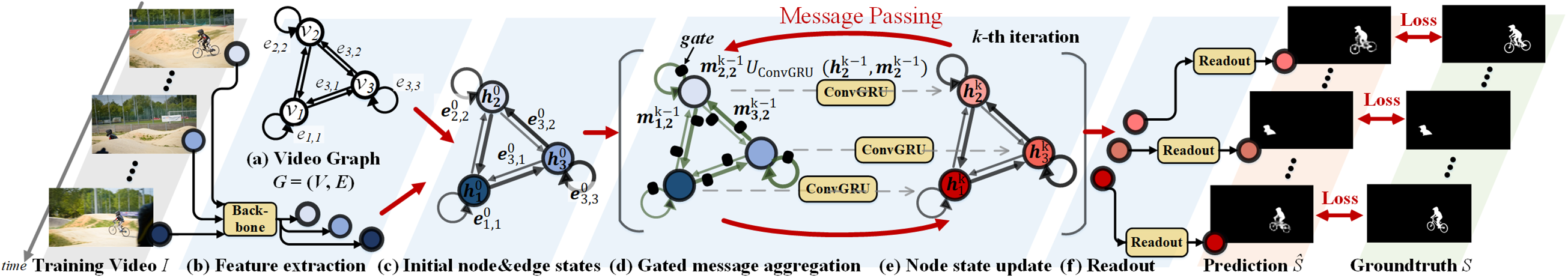}
\vspace{-20pt}
\caption{\small Our AGNN based ZVOS model during the training phase (see \S\ref{sec:agnn} and \S\ref{sec:detial}). \small Zoom in for details. }
\label{fig:model}
\vspace{-12pt}
\end{figure*}

\vspace*{-4pt}
\section{Our Algorithm}
\vspace*{-4pt}	
Before elaborating on our proposed AGNN (\S\ref{sec:agnn}), we first give a brief introduction to generic formulations of GNN models (\S\ref{sec:forgnn}). Finally, in \S\ref{sec:detial}, we provide detailed information on our network architecture.
	\vspace*{-3pt}
\subsection{General Formulations of GNNs}\label{sec:forgnn}
	\vspace*{-3pt}	
Based on deep neural networks and graph theory,  GNNs are powerful for collectively aggregating information from data represented in graph domains~\cite{scarselli2009graph,gilmer2017neural}. Specifically, a GNN model is defined according to a graph $\mathcal{G}\!=\!(\mathcal{V},\mathcal{E})$. Each node $v_i\!\in\!\mathcal{V}$ takes a unique value from $\{1,\dots,|\mathcal{V}|\}$, is associated with an initial \textit{node representation}
(or \textit{node state} or \textit{node embedding}) $\mathbf{v}_i$. Each edge $e_{i,j}\!\in\!\mathcal{E}$ is a pair $e_{i,j}\!=\!(v_i,v_j)\!\in\!|\mathcal{V}|\!\times\!|\mathcal{V}|$, with an \textit{edge representation} $\mathbf{e}_{i,j}$. For each node $v_i$, we learn an updated node representation $\mathbf{h}_i$ through aggregating representations
of its neighbors. Here $\mathbf{h}_i$ is used to produce an output $\mathbf{o}_i$, \ie, a node label.
More specifically, GNNs map graph $\mathcal{G}$ to the node outputs $\{\mathbf{o}_i\}_{i=1}^{|\mathcal{V}|}$ through two phases. First, a parametric \textit{message passing phase} runs for $K$ steps, which recursively propagates messages and updates node representations. At the $k$-th iteration, for each node $v_i$, we update its state according to its received message $\mathbf{m}^k_i$ (\ie, summarized information from its neighbors $\mathcal{N}_i$) and its previous state $\mathbf{h}^{k-1}_i$:
	\vspace*{-4pt}	
\begin{equation}
\begin{aligned}
    \!\!\!\!\!\!\text{\small{message aggregation:}}~~\mathbf{m}^k_i &\!=\!\sum\nolimits_{v_j\in \mathcal{N}_i}\mathbf{m}^k_{j,i},\\
    &\!=\!\sum\nolimits_{v_j\in \mathcal{N}_i}\!{M}(\mathbf{h}^{k-1}_j\!, \mathbf{e}^{k-1}_{i,j}),\!\!\\
    \!\!\!\!\!\!\text{\small{node representation update:}}~~~\mathbf{h}_i^k &\!=\! U(\mathbf{h}_i^{k-1}\!, \mathbf{m}_i^k),\!\!
    \end{aligned}
	\label{intro1}
	\vspace*{-4pt}	
\end{equation}
where $\mathbf{h}_i^0\!=\!\mathbf{v}_i$, $M(\cdot)$ and $U(\cdot)$ are the \textit{message function} and \textit{state update function}, respectively.
After $k$ iterations of aggregation, $\mathbf{h}^k_i$ captures the relations within the $k$-hop neighborhood of node $v_i$.

Second, a \textit{readout phase} maps the node representation $\mathbf{h}_i^K$ of the final $K$-iteration to a node output, through a \textit{readout function} $R(\cdot)$:
	\vspace*{-4pt}	
\begin{equation}
    \text{\small readout:}~~~\mathbf{o}_i = R(\mathbf{h}_i^K).
	\label{intro2}
	\vspace*{-4pt}	
\end{equation}
The message function $M$, update function $U$, and readout function $R$ are all learned differentiable functions.

Next, we present our AGNN based ZVOS solution, which essentially extends traditional fully connected GNNs to (1) preserve spatial features; and (2) capture pair-wise relations (edges) via a differentiable attention mechanism.
	\vspace*{-2pt}	
\subsection{Attentive Graph Neural Network}\label{sec:agnn}
	\vspace*{-1pt}	
\noindent\textbf{Problem Definition and Notations.} Given a set of training samples and an unseen testing video $\mathcal{I} \!=\! \{{I}_i\!\in\!\mathbb{R}^{w\times h \times 3}\}_{i=1}^{N}$ with $N$ frames in total, the goal of ZVOS is to generate a corresponding sequence of binary segment masks: $\mathcal{S} \!=\! \{{S}_i\!\in\!\{0,1\}^{w\times h}\}_{i=1}^{N}$.
To achieve this, AGNN represents $\mathcal{I}$ as a directed graph $\mathcal{G}\!=\!(\mathcal{V},\mathcal{E})$, where node $v_i\!\in\!\mathcal{V}$ represents the $i$-th frame ${I}_i$, and edge $e_{i,j}\!=\!(v_i,v_j)\!\in\!\mathcal{E}$ indicates the relation from ${I}_i$ to ${I}_j$. To comprehensively capture the underlying relationships between video frames, we assume $\mathcal{G}$ is fully connected and includes self-connections at each node (see Fig.~\ref{fig:model} (a)). For clarity, we refer to $e_{i,i}$, which connects a node $v_i$ to itself, as a \textit{loop-edge}; and $e_{i,j}$, which connects two different nodes $v_i$ and $v_j$, as a \textit{line-edge}.

The core idea of our AGNN  is to perform $K$ message propagation iterations over $\mathcal{G}$ to efficiently mine rich and high-order relations within $\mathcal{I}$. This helps to better capture the video content from a global view and obtain more accurate foreground estimates. We then readout the segmentation predictions $\hat{\mathcal{S}}$ from the final node states $\{\mathbf{h}^K_i\}_{i=1}^{N}$. Next, we describe each component of our model in detail.

\begin{figure*}[t]
  \centering
      \includegraphics[width=1 \linewidth]{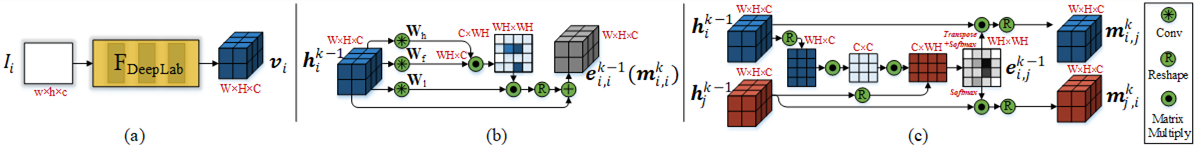}
\vspace{-22pt}
\caption{\small Detailed illustration of our (a) node embedding, (b) intra-attention based loop-edge embedding and corresponding loop-message generation, (c) inter-attention based straight-edge embedding and corresponding neighbor message generation. }
\label{fig:message}
\vspace{-12pt}
\end{figure*}


\noindent\textbf{FCN-Based Node Embedding.} We leverage DeepLabV3 \cite{DBLP:journals/corr/ChenPSA17}, a classical FCN based semantic segmentation architecture, to extract effective frame features, as node representations  (see Fig.~\ref{fig:model} (b) and Fig.~\ref{fig:message} (a)). For node $v_i$, its initial embedding $\mathbf{h}^0_i$ can be computed as:
	\vspace*{-4pt}
\begin{equation}
\mathbf{h}^0_i=\mathbf{v}_i=F_{\text{DeepLab}}(I_i)\in\!\mathbb{R}^{W\times H \times C},
	\vspace*{-4pt}
\end{equation}
where $\mathbf{h}^0_i$ is a 3D tensor feature with $W\!\times\!H$ spatial resolution and $C$ channels, which preserves spatial information as well as high-level semantic information.

\noindent\textbf{Intra-Attention Based Loop-Edge Embedding.} A loop-edge $e_{i,i}\!\in\!\mathcal{E}$ is a special edge that connects a node to itself. The loop-edge embedding $\mathbf{e}_{i,i}^k$ is used to capture the intra relations within node representation $\mathbf{h}_{i}^k$ (\ie, internal frame representation).
We formulate $\mathbf{e}_{i,i}^k$ as an \textit{intra-attention} mechanism~\cite{DBLP:conf/nips/VaswaniSPUJGKP17,wang2018non}, which has been proven complementary to convolutions and helpful for modeling long-range, multi-level dependencies across
image regions~\cite{zhang2018self}. In particular, the intra-attention calculates the response at a position by attending to all the positions within the same node embedding (see Fig.~\ref{fig:model} (c)  and Fig.~\ref{fig:message} (b)):
	\vspace*{-3pt}
\begin{equation}\small
\begin{aligned}
\!\!\!\!\mathbf{e}_{i,i}^k\! &=\!F_{\text{intra-att}}(\mathbf{h}_{i}^k) \in \mathbb{R}^{W\times H \times C}\\
&=\!\alpha~\text{softmax}\big((\mathbf{W}_{\!f} \!*\! \mathbf{h}_{i}^k) (\mathbf{W}_{\!h}  \!*\! \mathbf{h}_{i}^k)^\top \big)( \mathbf{W}_{\!l} \!*\! \mathbf{h}_{i}^k) \!+\! \mathbf{h}_{i}^k,\!\!
\label{self-attention}
\end{aligned}
	\vspace*{-3pt}
\end{equation}
where \!`$*$'\! represents the convolution operation, $\mathbf{W}$s indicate learnable convolution kernels, and $\alpha$ is a learnable scale parameter. Eq.~\ref{self-attention} makes the output element of each position in  $\mathbf{h}_{i}^k$ encode contextual information as well as its original information, thus enhancing the representability.

\noindent\textbf{Inter-Attention Based Line-Edge Embedding.}  A line-edge $e_{ij}\!\in\!\mathcal{E}$ connects two different nodes $v_i$ and $v_j$. The line-edge embedding $\mathbf{e}_{i,j}^k$ is used to mine the relation from node $v_i$ to $v_j$, in the node embedding space (see Fig.~\ref{fig:model} (b)).  Here we compute an \textit{inter-attention} mechanism~\cite{lu2016hierarchical} to capture the bi-directional relations between two nodes $v_i$ and $v_j$ (see Fig.~\ref{fig:model} (c) and Fig.~\ref{fig:message} (c)):
	\vspace*{-3pt}
\begin{equation}
\begin{aligned}
\!\!\mathbf{e}_{i,j}^k &\!=\! F_{\text{inter-att}}(\mathbf{h}_i^k, \mathbf{h}_j^k) \!=\! {\mathbf{h}^{k }_i} \mathbf{W}_c {\mathbf{h}_j^{k \top}} \in \mathbb{R}^{(W\!H) \times (W\!H)}, \\
\!\!\mathbf{e}_{j,i}^k &\!=\! F_{\text{inter-att}}(\mathbf{h}_j^k, \mathbf{h}_i^k) \!=\! {\mathbf{h}_j^{k }} \mathbf{W}_c^{\top}{\mathbf{h}_i^{k\top}}\!\in \mathbb{R}^{(W\!H) \times (W\!H)},
\end{aligned}
\label{edge}
\vspace*{-2pt}
\end{equation}
where $\mathbf{e}_{i,j}^k={\mathbf{e}_{j,i}^{k\top}}$. $\mathbf{e}_{i,j}^k$  indicates the outgoing edge feature and $\mathbf{e}_{j,i}^k$ the incoming one, for node $v_i$. $\mathbf{W}_{\!c\!} \in\!\mathbb{R}^{C\times C\!}$ indicates a learnable weight matrix. 
$\mathbf{h}^k_{j\!} \!\in\!\mathbb{R}^{{(W\!H) \times C}\!\!}$ and $\mathbf{h}^k_i \!\in\! \mathbb{R}^{{(W\!H)} \times C\!}$  are flattened into matrix representations. Each element in $\mathbf{e}^k_{i,j}$ reflects the similarity between each row of $\mathbf{h}^{k}_{i}$ and each column of $\mathbf{h}^{k\top}_{j}$. As a result, $\mathbf{e}^k_{i,j}$ can be viewed as the \textit{importance} of node $v_i$'s embedding to $v_j$, and vice versa. By attending to each node pair, $\mathbf{e}^k_{i,j}$ explores their joint representations in the node embedding space.   

\noindent\textbf{Gated Message Aggregation.}
In our AGNN, for the message passed in the self-loop, we view the loop-edge embedding $\mathbf{e}_{i,j}^{k-1}$ itself as a message (see Fig.~\ref{fig:message} (b)), since it already contains the contextual and original node information (see Eq.~\ref{self-attention}):
	\vspace*{-3pt}
\begin{equation}
\begin{aligned}
\mathbf{m}_{i,i}^k&\!=\mathbf{e}^{k-1}_{i,i}  \!\in\!  \mathbb{R}^{ W\times H \times C}.
\label{message1}
\end{aligned}
	\vspace*{-3pt}
\end{equation}
For the message $\mathbf{m}_{j,i}$ passed from node $v_j$ to $v_i$ (see Fig.~\ref{fig:message} (c)), we have:
	\vspace*{-2pt}
\begin{equation}\small
\begin{aligned}
\!\!\mathbf{m}_{j,i}^k\!=\!M(\mathbf{h}_j^{k-1}\!, \mathbf{e}^{k-1}_{i,j}) \!=\! \text{softmax}(\mathbf{e}^{k-1}_{i,j}) \mathbf{h}_j^{k-1 }   \!\in\!  \mathbb{R}^{ (W\!H) \times C},
\label{message2}
\end{aligned}
	\vspace*{-2pt}
\end{equation}
where softmax($\cdot$) normalizes each row of the input. Thus, each row (position) of $\mathbf{m}_{j,i}^k$ is a weighted combination of each row (position) of $\mathbf{h}_j^{k-1}$, where the weights come from the corresponding column of $\mathbf{e}^{k-1}_{i,j}$.   In this way, the message function $M(\cdot)$ assigns its edge-weighted feature (\ie, message) to the neighbor nodes~\cite{velickovic2017graph}. Then, $\mathbf{m}^k_{j,i}$ is reshaped back to a 3D tensor with a size of $W\!\times\!H\!\times\!C$.

In addition, because some nodes are noisy due to camera shift or out-of-view, their messages may be useless or even harmful. We apply a learnable gate $G(\cdot)$ to measure the confidence of a message $\mathbf{m}_{j,i}$:
	\vspace*{-2pt}
\begin{equation}
\begin{aligned}
\mathbf{g}_{j,i}^k\!=\!G(\mathbf{m}_{j,i}^k)\!=\!\sigma \big(F_{\text{GAP}} (\mathbf{W}_g\!*\!\mathbf{m}_{j,i}^k\!+\!b_g) \big) \!\in\!  [0,1]^{C},
\end{aligned}	
\label{equ:gate1}
	\vspace*{-2pt}
\end{equation}
where $F_{\text{GAP}}(\cdot)$ indicates the use of global average pooling to generate channel-wise responses, $\sigma$ is the logistic sigmoid function $\sigma(x)\!=\!1/(1\!+\!\exp(-x))$, and $\mathbf{W}_g$ and $b_g$ are the trainable convolution kernel and bias.



Following Eq.~\ref{intro1}, we collect the messages from the neighbors and self-loop via gated summarization (see Fig.~\ref{fig:model} (d)):
	\vspace*{-7pt}
\begin{equation}
\mathbf{m}_i^k = \sum\nolimits_{v_j \in\mathcal{V}}\mathbf{g}_{j,i}^k\star\mathbf{m}_{j,i}^k \in  \mathbb{R}^{W\times H\times C},
\label{aggregation}
	\vspace*{-3pt}
\end{equation}
where `$\star$' denotes the channel-wise Hadamard product. Here, the gate mechanism is used to filter out irrelevant information from noisy frames. See \S\ref{sec:ablation} for a quantitative study of this design.


\noindent\textbf{ConvGRU based Node-State Update.}
In step $k$, after aggregating all the information from the neighbor nodes and itself (Eq.~\ref{aggregation}), $v_i$ gets a new state $\mathbf{h}_i^{k}$ by taking into account its prior state $\mathbf{h}_i^{k-1}$ and its received message $\mathbf{m}_i^k$. To preserve the spatial information conveyed in $\mathbf{h}_i^{k-1}$ and $\mathbf{m}_i^k$, we leverage ConvGRU~\cite{ballas2015delving} to update the node state (Fig.~\ref{fig:model} (e)):
	\vspace*{-6pt}
\begin{equation}
\begin{aligned}
\mathbf{h}_i^k = U_{\text{ConvGRU}}(\mathbf{h}_i^{k-1}, \mathbf{m}_i^k) \!\in\!  \mathbb{R}^{ W\times H \times C}.
\label{convGRU}
\end{aligned}
	\vspace*{-1pt}
\end{equation}
ConvGRU is proposed as a convolutional counterpart to previous fully connected
GRU~\cite{cho2014learning}, and introduces convolution operation into input-to-state
and state-to-state transitions.

\noindent\textbf{Readout Function.}
After $K$ message passing iterations, we obtain the final state $\mathbf{h}_i^K$ for each node $v_i$.
Finally, in the readout phase, we get a segmentation prediction map $\hat{S}\!\in\![0,1]^{W\times H}$ from $\mathbf{h}_i^K$ through a readout function $R(\cdot)$  (see Fig.~\ref{fig:model} (f)). Slightly different from Eq.~\ref{intro2}, we concatenate the final node state $\mathbf{h}_i^K$ and the original node feature $\mathbf{v}_i$ (\ie, $\mathbf{h}_i^0$) together and feed the combined feature into $R(\cdot)$:
	\vspace*{-4pt}
\begin{equation}
\hat{S}_i = R_{\text{FCN}}([\mathbf{h}_i^K, \mathbf{v}_i]) \!\in\!  [0,1]^{ W\times H }.
\label{readout}
	\vspace*{-3pt}
\end{equation}
Again, to preserve spatial information, the readout function is implemented as a small FCN network, which has three convolution layers with a sigmoid function to normalize the prediction to $[0,1]$.

The convolution operations in the intra-attention (Eq.~\ref{self-attention}) and update function (Eq.~\ref{convGRU}) are realized with $1\!\times\!1$ convolutional layers. The readout function (Eq.~\ref{readout}) consists of two $3\!\times\!3$ convolutional layers cascaded by a $1\!\times\!1$ convolutional layer. As a message passing based GNN model, these functions share weights among all the nodes. Moreover, all the above functions are carefully designed to avoid disturbing spatial information, which is essential for ZVOS since it is  a pixel-wise prediction task.

\subsection{Detailed Network Architecture}\label{sec:detial}
	\vspace*{-2pt}	
Our whole model is  end-to-end trainable, as all the functions in AGNN are parameterized by neural networks. We use the first five convolution blocks of DeepLabV3~\cite{DBLP:journals/corr/ChenPSA17} as our backbone for feature extraction. For an input video $\mathcal{I}$, each frame $I_i$ (with a resolution of $473\!\times\!473$) is represented as a node $v_i$ in the video graph $\mathcal{G}$ and associated with an initial node state $\textbf{v}_i\!=\! \textbf{h}^0_i\!\in\!\mathbb{R}^{60\times 60 \times 256}$. Then, after a total of $K$ message passing iterations, for each node $v_i$, we use the readout function in Eq.~\ref{readout} to obtain a corresponding segmentation prediction map $\hat{S}\!\in\![0,1]^{60\times 60}$. More details on the training and testing phases  are provided as follows.

\begin{table*}[t]
	\centering
	\begin{threeparttable}
		\resizebox{0.99\textwidth}{!}{
			\setlength\tabcolsep{3pt}
			\renewcommand\arraystretch{1.0}
			\begin{tabular}{|lc||ccccccccccccc|c|}
				\hline\thickhline
				&Method&KEY~\cite{lee2011key} &MSG~\cite{DBLP:conf/iccv/OchsB11} &NLC~\cite{DBLP:conf/bmvc/FaktorI14} &CUT~\cite{DBLP:conf/iccv/KeuperAB15}&FST~\cite{DBLP:conf/iccv/PapazoglouF13} &SFL~\cite{cheng2017segflow}&MP~\cite{DBLP:conf/cvpr/TokmakovAS17} &FSEG~\cite{jain2017fusionseg}  & LVO~\cite{DBLP:conf/iccv/TokmakovAS17} &ARP~\cite{DBLP:conf/cvpr/KohK17}   &{PDB}~\cite{Song_2018_ECCV}
				& MOA~\cite{siam2018video} &AGS~\cite{wang2019learning}&\textbf{AGNN}\\ 
				\hline
				\hline
				\multirow{3}{*}{ $\mathcal{J}$\!\!} & Mean $\uparrow$    &  49.8& 53.3& 55.1& 55.2 & 55.8& 67.4& 70.0 & 70.7 & 75.9 &76.2  & {77.2}
&77.2 &\multicolumn{1}{>{\columncolor{mygray}}c}{79.7}& \multicolumn{1}{>{\columncolor{mygray1}}c|}{\textbf{80.7}} \\
				& Recall $\uparrow$   & 59.1& 61.6& 55.8 &57.5 &64.9 &81.4 & 85.0 & 83.0&89.1& \multicolumn{1}{>{\columncolor{mygray}}c}{91.1}  &90.1
&87.8&\multicolumn{1}{>{\columncolor{mygray}}c}{91.1}&\multicolumn{1}{>{\columncolor{mygray1}}c|}{\textbf{94.0}}  \\ 
				
				& Decay~$\downarrow$    & 14.1& 2.4 & 12.6 & 2.2 & \multicolumn{1}{>{\columncolor{mygray1}}c}{\textbf{0.0}}&6.2 &1.3 & 1.5 &\multicolumn{1}{>{\columncolor{mygray1}}c}{\textbf{0.0}} & 7.0&0.9
&5.0&1.9&\multicolumn{1}{>{\columncolor{mygray}}c|}{{0.03}} \\\hline 
				\multirow{3}{*}{ $\mathcal{F}$\!\!} & Mean $\uparrow$     &42.7 & 50.8& 52.3&55.2 &51.1 &66.7 &65.9 & 65.3& 72.1& 70.6  & 74.5
&\multicolumn{1}{>{\columncolor{mygray}}c}{77.4}&\multicolumn{1}{>{\columncolor{mygray}}c}{77.4}& \multicolumn{1}{>{\columncolor{mygray1}}c|}{\textbf{79.1}} \\ 
				
				& Recall $\uparrow$  & 37.5&60.0 & 61.0& 51.9&51.6 &77.1 &79.2 &73.8 &83.4 & 83.5&  {84.4}
&{84.4}&\multicolumn{1}{>{\columncolor{mygray}}c}{85.8}& \multicolumn{1}{>{\columncolor{mygray1}}c|}{\textbf{90.5}}\\ 
				& Decay~$\downarrow$     &10.6 & 5.1& 11.4 &3.4 &2.9 & 5.1 &2.5&1.8 & 1.3 &7.9 &\multicolumn{1}{>{\columncolor{mygray1}}c}{\textbf{-0.2}}
&3.3&1.6&\multicolumn{1}{>{\columncolor{mygray}}c|}{0.03} 
				\\\hline
				$\mathcal{T}\!\!$  & Mean~$\downarrow$   & {26.9}& 30.2& 42.5 &27.7  &36.6 &28.2 & 57.2&32.8 &\multicolumn{1}{>{\columncolor{mygray1}}c}{\textbf{26.5}} &39.3& 29.1
&27.9&\multicolumn{1}{>{\columncolor{mygray}}c}{26.7}&33.7		\\\hline \thickhline 
			\end{tabular}
		}
	\end{threeparttable}
	\vspace*{-8pt}
	\caption{\small Quantitative results on the validation set of DAVIS$_{16}$~\cite{perazzi2016benchmark} (\S\ref{sec:exqUVOS}). 
The scores are borrowed from the public leaderboard\textsuperscript{\ref{web}}.
(The best scores are marked in \textbf{bold}. The best two entries in each row are marked in gray. These notes are the same to other tables.  ) }
	\label{davis}
	\vspace*{-10pt}	
\end{table*}

\begin{table*}[t]\small
	\centering
	\resizebox{0.9\textwidth}{!}{
		\setlength\tabcolsep{4pt}
		\renewcommand\arraystretch{1}
		\begin{tabular}{|c||cccccccccc|c|}
			\hline\thickhline
			& Airplane (6) & Bird (6) & Boat (15)  & Car (7) & Cat (16)& Cow (20) & Dog (27) &Horse (14)& Motorbike (10)& Train (5) & Avg. \\
			\hline
			\hline
			FST~\cite{DBLP:conf/iccv/PapazoglouF13}& 70.9 &70.6& 42.5&65.2  &52.1&44.5&65.3 &53.5&44.2 &29.6 &53.8  \\		
			COSEG~\cite{tsai2016semantic}&69.3   &76.0 &53.5  &70.4 & 66.8& 49.0 & 47.5& 55.7& 39.5& 53.4& 58.1\\
			ARP~\cite{DBLP:conf/cvpr/KohK17} &73.6 & 56.1&57.8 &33.9& 30.5& 41.8& 36.8 &44.3& 48.9& 39.2&46.2\\
			LVO~\cite{DBLP:conf/iccv/TokmakovAS17} &86.2 &81.0&68.5 &69.3& 58.8&68.5&61.7 &53.9& 60.8& 66.3&67.5\\
			PDB~\cite{Song_2018_ECCV} &78.0  &80.0 &58.9 &76.5 &63.0&64.1&70.1 &67.6&58.3&35.2&65.4\\
			FSEG~\cite{jain2017fusionseg}  &{81.7}  &63.8 & 72.3&74.9&68.4&68.0 &69.4 & 60.4&62.7&62.2&68.4\\
			SFL~\cite{cheng2017segflow}&65.6  &65.4 &59.9&64.0 &58.9&51.1& 54.1 &64.8& 52.6& 34.0&57.0\\
AGS~\cite{wang2019learning} &87.7&76.7&72.2&78.6&69.2&64.6&73.3&64.4&62.1&48.2&\multicolumn{1}{>{\columncolor{mygray}}c|}{69.7}\\
\hline
			\textbf{AGNN} &81.1  &75.9 &{70.7} &78.1 &67.9&69.7&77.4 &67.3&68.3&47.8&\multicolumn{1}{>{\columncolor{mygray1}}c|}{\textbf{70.8}}\\
			\hline \thickhline
		\end{tabular}
	}
\vspace{-7pt}
\caption{\small Quantitative performance of each category on Youtube-Objects~\cite{DBLP:conf/cvpr/PrestLCSF12} (\S\ref{sec:exqUVOS}) with mean $\mathcal{J}$. We show the average performance for each of the 10 categories, and the final row shows an average over all the videos.
	}
	\label{Youtube-Objects}
	\vspace*{-15pt}	
\end{table*}


\noindent\textbf{Training Phase.} As we operate on batches of a certain size (which is allowed to vary, depending on the GPU memory size), we leverage a random sampling strategy to train AGNN. Specifically, we split each training video $\mathcal{I}$ with a total of $N$ frames into $N'$ segments ($N'\!\leq\!N$) and randomly select one frame from each segment. Then we feed the $N'$ sampled frames into a batch and train AGNN. Thus the relationships among all the $N'$ sampling frames in each batch are represented using an $N'$-node graph. Such a sampling strategy provides robustness to variations and enables the network to fully exploit all frames. The diversity among the samples enables our model to better capture the underlying relationships and improve its generalizability. Let us denote the ground-truth segmentation mask and predicted foreground map for a training frame $I_i$ as $S\!\in\!\{0,1\}^{60\times 60}$ and $\hat{S}\!\in\![0,1]^{60\times 60}$. Our model is trained through the weighted binary cross entropy loss  (see Fig.~\ref{fig:model}):
	\vspace*{-3pt}
\begin{equation}\small
\!\!\!\!\mathcal{L}(S\!, \hat{S})\!=\!-\!\!\sum^{W\times H}\nolimits_{x} \!(1\!-\!{\eta}) {S}_{x\!}\log({\hat{S}_{x}})\! +\!  {\eta}(1\!-\!{S}_{x})\log(1\!-\!{\hat{S}_{x}}),\!\!
\label{focal_loss}	
	\vspace*{-0pt}
\end{equation}
where $\eta$ indicates the foreground-background pixel number ratio in $S$.
It is worth mentioning that, as AGNN handles multiple video frames at the same time, it leads to a remarkably efficient training data augmentation strategy, as the combination candidates are numerous. In our experiments, during training, we randomly select 2 videos from the training video set and sample 3 frames ($N'\!=\!3$) per video, due to the computation limitation. In addition, we set the total number of iterations as $K\!=\!3$. Quantitative experimental settings can be found in \S\ref{sec:ablation}.

\noindent\textbf{Testing Phase.} After training, we can apply the learned AGNN model to perform per-pixel object prediction over unseen videos. For an input test video $\mathcal{I}$ with $N$ frames (with $473\!\times\!473$ resolution), we split $\mathcal{I}$ into $T$ subsets: $\{\mathcal{I}_1,\mathcal{I}_2,\dots,\mathcal{I}_T\}$, where $T\!=\!N/N'$. Each subset contains $N'$ frames with an interval of $T$ frames: $\mathcal{I}_t\!=\!\{I_t,I_{t+T},\dots,I_{N-T+t}\}$. 
Then we feed each subset into AGNN to obtain the segmentation maps of all the frames in the subset. In practice, we set $N'\!=\!5$ during testing. We quantitatively study this setting in \S\ref{sec:ablation}. As our AGNN does not require time-consuming optical flow computation and processes $N'$ frames in one feed-forward propagation, it achieves a fast speed of $0.28s$ per frame.  Following the widely used protocol~\cite{DBLP:conf/iccv/TokmakovAS17,DBLP:conf/cvpr/TokmakovAS17,Song_2018_ECCV}, we apply CRF as a post-processing step, which takes about $0.50s$ per frame. More implementation details can be found in \S\ref{sec:exUVOSimpl}.


\vspace*{-4pt}
\section{Experiments}
	\vspace*{-2pt}	
We first report performance on the main task: {\color {black}unsupervised video object segmentation (\S\ref{sec:exUVOS})}. Then, in \S\ref{sec:exIOCS}, to further demonstrate the advantages of our AGNN model, we test it on an additional task: image object co-segmentation. Finally, we conduct an ablation study in \S\ref{sec:ablation}.
\vspace*{-3pt}
\subsection{Main Task: ZVOS}\label{sec:exUVOS}
\vspace*{-3pt}
\subsubsection{Experimental Setup}\label{sec:exUVOSimpl}
\vspace*{-6pt}
\noindent\textbf{Datasets and Metrics:}~ {\color{black}We use two well-known datasets:} \vspace*{-1pt}
\vspace*{-1pt}
\begin{itemize}[leftmargin=*]
	\setlength{\itemsep}{0pt}
	\setlength{\parsep}{-2pt}
	\setlength{\parskip}{-0pt}
	\setlength{\leftmargin}{-15pt}
	\vspace{-5pt}
	\item \textbf{DAVIS$_{16}$~\cite{perazzi2016benchmark}} is a challenging video object segmentation dataset which consists of 50 videos in total (30 for training and 20 for val) with pixel-wise annotations for every frame. Three evaluation criteria are used in this dataset, \ie, region similarity (Intersection-over-Union) $\mathcal{J}$, boundary accuracy $\mathcal{F}$, and time stability $\mathcal{T}$.
	
	
	\item \textbf{Youtube-Objects~\cite{DBLP:conf/cvpr/PrestLCSF12}} comprises 126 video sequences which belong to 10 object categories  and contain more than 20,000 frames in total. Following its protocol, we use $\mathcal{J}$ to measure the segmentation performance.
	\item \textbf{DAVIS$_{17}$~\cite{pont20172017}} consists of 60 videos in the training set, 30 videos in the validation set and 30 videos in the test-dev set. Different from DAVIS2016 and Youtube-Objects, which only focus on object-level video object segmentation, DAVIS$_{17}$ provides instance-level annotations.
\vspace*{-3pt}
\end{itemize}

\noindent\textbf{Implementation Details:}~Following~\cite{DBLP:conf/cvpr/PerazziKBSS17,Song_2018_ECCV}, both static data from image salient object segmentation datasets, {MSRA10K}~\cite{cheng2015global}, {DUT}~\cite{DBLP:conf/cvpr/YangZLRY13}, and video data from the training set of DAVIS$_{16}$  are iteratively used to train our model. In a `static-image' iteration, we randomly sample 6 images from the static training data to train our backbone network (DeepLabV3) to extract more discriminative foreground  features. To train the backbone network,  a $1\!\times\!1$ convolution layer with \textit{sigmoid} function is appended as an intermediate output layer, which can access the static image supervision signal. This is followed by a `dynamic-video' iteration, in which we use the sampling strategy described in \S\ref{sec:detial} to sample 6 video frames to train our whole AGNN model.
The `static-image' and `dynamic-video' iterations are executed alternately. To apply the trained AGNN model on DAVIS$_{17}$, we first use category agnostic mask-RCNN~\cite{he2017mask} to generate instance-level object proposals for each frame. Then, we run AGNN on the whole video and generate a coarse mask for the primary objects in each frame. Then the object-level masks are used to filter out the proposals from the background and highlight the foreground proposals. Through combining an instance bounding proposals and coarse masks, we obtain the instance-level mask for each primary object. Finally, to connect multiple instances across different frames, we use overlap ratio and optical flow as an association metric~\cite{luiten2018premvos} to match different instance-level masks.

\begin{figure*}[t]
	\centering
	\includegraphics[width=.98\textwidth]{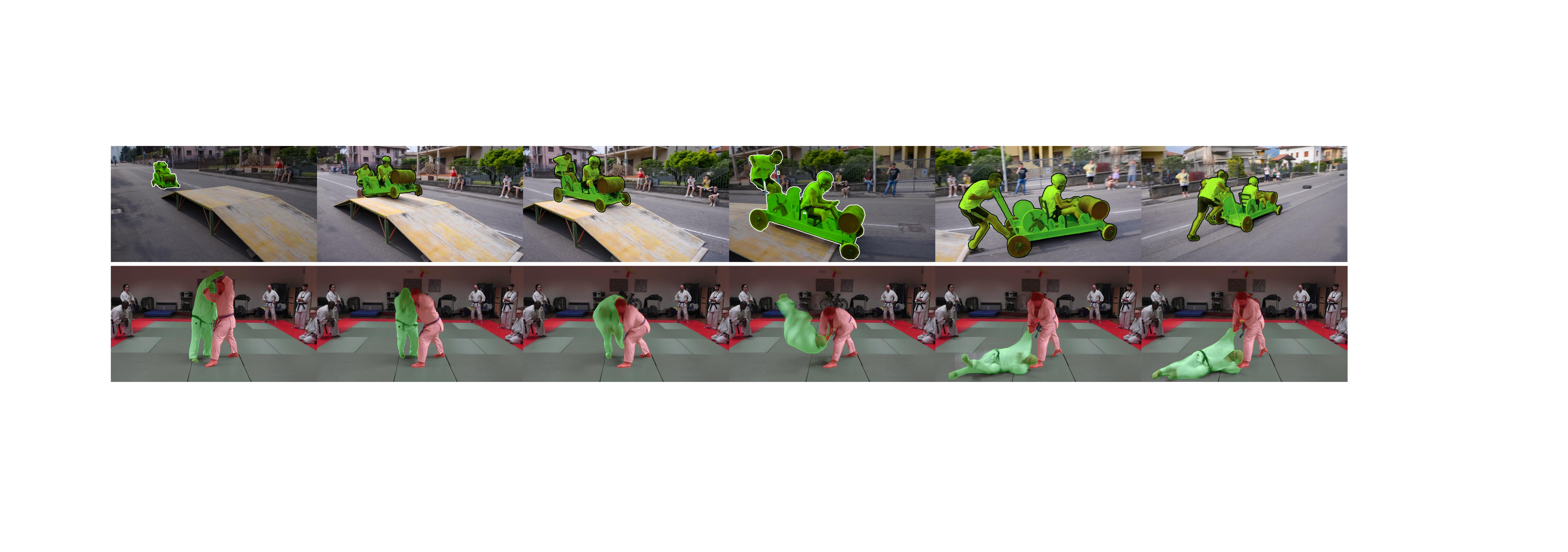}
	\vspace*{-9pt}	
	\caption{\small Qualitative results on two example videos (top: \textit{soapbox}, bottom: \textit{judo}) from the DAVIS$_{16}$ val set and DAVIS$_{17}$ test-dev set, respectively (see \S\ref{sec:exvisualUVOS}). }
	\label{fig:uvos}	
	\vspace*{-18pt}	
\end{figure*}

\vspace*{-12pt}
\subsubsection{Quantitative Performance}\label{sec:exqUVOS}
\vspace*{-4pt}
\noindent\textbf{Val-set of DAVIS$_{16}$.}
We compare the proposed AGNN with the top ZVOS methods from the DAVIS$_{16}$ benchmark\footnote{\scriptsize{\url{https://davischallenge.org/davis2016/soa_compare.html}, deadline: Mar. 2019 }\label{web}}~\cite{perazzi2016benchmark}. 
 Table~\ref{davis} shows the detailed results. We can see that our AGNN outperforms the best reported results (\ie, AGS~\cite{wang2019learning}) on DAVIS$_{16}$ benchmark by a significant margin in terms of mean $\mathcal{J}$ (80.7 \textit{vs} 79.7) and $\mathcal{F}$ (79.1 \textit{vs} 77.4). 
 Compared to  PDB~\cite{Song_2018_ECCV}, which uses the same training protocol and training datasets, our AGNN yields significant performance gains of 3.5$\%$ and 4.6$\%$ in terms of mean $\mathcal{J}$ and mean $\mathcal{F}$, respectively.

\noindent\textbf{Youtube-Objects.} Table~\ref{Youtube-Objects} gives the detailed per-category performance and average results on Youtube-Objects. As can be seen, our AGNN performs favorably according to mean $\mathcal{J}$ criterion. Furthermore, unlike other methods whose performance fluctuates across categories, AGNN mains a stable performance.
This further proves its robustness and generalizability.

\noindent\textbf{Test-dev set of DAVIS$_{17}$.} In Table~\ref{Davis2017} we report the performance comparison with the recent instance-level ZVOS method, RVOS~\cite{ventura2019rvos}, on the
DAVIS$_{17}$ test-dev set. We can find that AGNN significantly outperforms RVOS over most evaluation criteria. 
\vspace*{-10pt}
\begin{table}[t]
	\centering
	\resizebox{0.49\textwidth}{!}{
		\setlength\tabcolsep{2pt}
		\renewcommand\arraystretch{1.0}
		\begin{tabular}{|c||ccc|ccc|c|}
			\hline \thickhline
			\multirow{2}{*}{Method} &\multicolumn{3}{c|}{$\mathcal{J}$} &\multicolumn{3}{c|}{$\mathcal{F}$}& \multirow{2}{*}{$\mathcal{J}$\&$\mathcal{F}$ Mean $\uparrow$}\!\\
 &Mean $\uparrow$  &Recall $\uparrow$ &Decay $\downarrow$&Mean $\uparrow$  &Recall $\uparrow$ &Decay $\downarrow$& \\
\hline
\hline			
			RVOS~\cite{ventura2019rvos}  & {39.0} & 42.8&\multicolumn{1}{>{\columncolor{mygray1}}c|}{\textbf{0.50}}&48.3 &49.6&\multicolumn{1}{>{\columncolor{mygray1}}c|} {\textbf{-0.01}} &{43.7}\\
			\textbf{AGNN} & \multicolumn{1}{>{\columncolor{mygray1}}c}{58.9} &  \multicolumn{1}{>{\columncolor{mygray1}}c}{\textbf{65.7}}&11.7& \multicolumn{1}{>{\columncolor{mygray1}}c}{\textbf{63.2}}&  \multicolumn{1}{>{\columncolor{mygray1}}c}{\textbf{67.1}} &  {14.3} & \multicolumn{1}{>{\columncolor{mygray1}}c|}{\textbf{61.1}} \\
			\hline \thickhline
	\end{tabular}}	
	\vspace*{-8pt}
	\caption{\small Quantitative results on the DAVIS$_{17}$ test-dev set~\cite{pont20172017}. 
}
	\vspace*{-15pt}
\label{Davis2017}	
\end{table}

\vspace*{-2pt}
\subsubsection{Qualitative Performance}\label{sec:exvisualUVOS}
\vspace*{-4pt}
Fig.~\ref{fig:uvos} depicts visual results for the proposed AGNN on two challenging video sequences \textit{soapbox} and \textit{judo} of DAVIS$_{16}$ and DAVIS$_{17}$, respectively.  
For \textit{soapbox}, the primary objects undergo huge scale variation, deformation and view changes, but our AGNN still generates accurate foreground segments. Our AGNN also handles \textit{judo} well, although the different foreground instances suffer from similar appearance and rapid motions.
\vspace*{-2pt}
\subsection{Additional Task: IOCS}\label{sec:exIOCS}
\vspace*{-2pt}
Our AGNN model can be viewed as a framework for capturing high-order relations among images (or frames). To demonstrate its  generalizability, we extend  AGNN for IOCS task. Rather than  extracting the foreground objects across multiple relatively similar video frames in videos, IOCS needs to infer the common objects from a group of semantically related images.
\vspace*{-10pt}
\subsubsection{Experimental Setup}
\vspace*{-4pt}
\noindent\textbf{Datasets and Metrics:} We perform experiments on two well-known IOCS datasets:
\begin{itemize}[leftmargin=*]
	\setlength{\itemsep}{0pt}
	\setlength{\parsep}{-2pt}
	\setlength{\parskip}{-0pt}
	\setlength{\leftmargin}{-15pt}
	\vspace{-4pt}
	\item \textbf{PASCAL VOC~\cite{pascal-voc-2012}} has 1,464 training images and 1,449 validation images. Following~\cite{DBLP:journals/corr/abs-1804-06423}, we split the validation set into 724 validation and 725 test images, and use mean $\mathcal{J}$ as the performance measure.
	
	\item \textbf{Internet~\cite{Rubinstein_2013_CVPR}}  contains 1,306 car, 879 horse, and 561 airplane images.  Following~\cite{chen2018semantic,quan2016object}, we measure the performance on a subset of Internet (100 images per class are sampled) with mean $\mathcal{J}$.
\vspace*{-2pt}
\end{itemize}

\begin{table}
	\centering
	\resizebox{0.48\textwidth}{!}{
		\setlength\tabcolsep{6pt}
		\renewcommand\arraystretch{1}
		\begin{tabular}{|c||ccc|c|}
			\hline\thickhline
			Method &GO-FMR~\cite{quan2016object}&FCNs~\cite{long2015fully} & CA~\cite{chen2018semantic} &\textbf{AGNN}\\
\hline
Mean $\mathcal{J}$  $\uparrow$ & 52.0& 55.21 &59.24& \multicolumn{1}{>{\columncolor{mygray1}}c|}{\textbf{60.78}} \\\hline
\hline
Method&FCA~\cite{chen2018semantic}&CSA~\cite{chen2018semantic} & DOCS~\cite{DBLP:journals/corr/abs-1804-06423} &\textbf{AGNN}\\	
\hline
Mean $\mathcal{J}$  $\uparrow$&59.41&\multicolumn{1}{>{\columncolor{mygray}}c}{59.76}& 57.82&\multicolumn{1}{>{\columncolor{mygray1}}c|}{\textbf{60.78}} \\
			\hline \thickhline
	\end{tabular}}
	\vspace*{-8pt}	
	\caption{\small Quantitative performance on PASCAL VOC~\cite{pascal-voc-2012} with mean $\mathcal{J}$. We show the average performance for 20 categories averaged over all the images. See \S\ref{sec:exqIOCS} for detailed analyses. 
	}
	\label{pascal}
	\vspace*{-16pt}	
\end{table}
		
\begin{table*}[t]\small
	\centering
	\resizebox{0.86\textwidth}{!}{
		\setlength\tabcolsep{2pt}
		\renewcommand\arraystretch{1}
		\begin{tabular}{|c||cccccccccc|c|}
			\hline\thickhline
			Method &DC~\cite{joulin2010discriminative}&Internet~\cite{Rubinstein_2013_CVPR}&TDK~\cite{chen2014enriching}& GO-FMR~\cite{quan2016object} &DDCRF~\cite{yuan2017deep} & CA~\cite{chen2018semantic}&FCA~\cite{chen2018semantic}&CSA~\cite{chen2018semantic} & DOCS~\cite{DBLP:journals/corr/abs-1804-06423} &CoA~\cite{hsu2018co} &\textbf{AGNN}\\ \hline
			\hline
			Car& 37.1&64.4&64.9&66.8&72.0&80.0&76.9&79.9&\multicolumn{1}{>{\columncolor{mygray}}c}{82.7}& 82.0& \multicolumn{1}{>{\columncolor{mygray1}}c|}{\textbf{84.0}} \\
			Horse& 30.1&51.6&33.4&58.1&65.0&67.3&69.1&\multicolumn{1}{>{\columncolor{mygray}}c}{71.4}&64.6& 61.0& \multicolumn{1}{>{\columncolor{mygray1}}c|}{\textbf{72.6}} \\
			Airplane & 15.3&57.3&46.2&60.4&67.7&72.8&70.6&\multicolumn{1}{>{\columncolor{mygray}}c}{73.1}&70.3& 67.0& \multicolumn{1}{>{\columncolor{mygray1}}c|}{\textbf{76.1}} \\
			\hline
			Avg.  &27.5& 57.3 &46.2&60.4&  67.7& 70.3& 72.8&70.6 &\multicolumn{1}{>{\columncolor{mygray}}c}{73.1}& 67.7& \multicolumn{1}{>{\columncolor{mygray1}}c|}{\textbf{77.6}}  \\		
			\hline \thickhline
	\end{tabular}}
	\vspace*{-8pt}	
	\caption{\small Quantitative results on Internet~\cite{Rubinstein_2013_CVPR} with mean $\mathcal{J}$ (\S\ref{sec:exqIOCS}). We show the per-class performance and an overall average.   
	}
	\label{Internet}
	\vspace*{-10pt}	
\end{table*}

\begin{figure*}[t]
	\centering
	~\includegraphics[width=.95\textwidth]{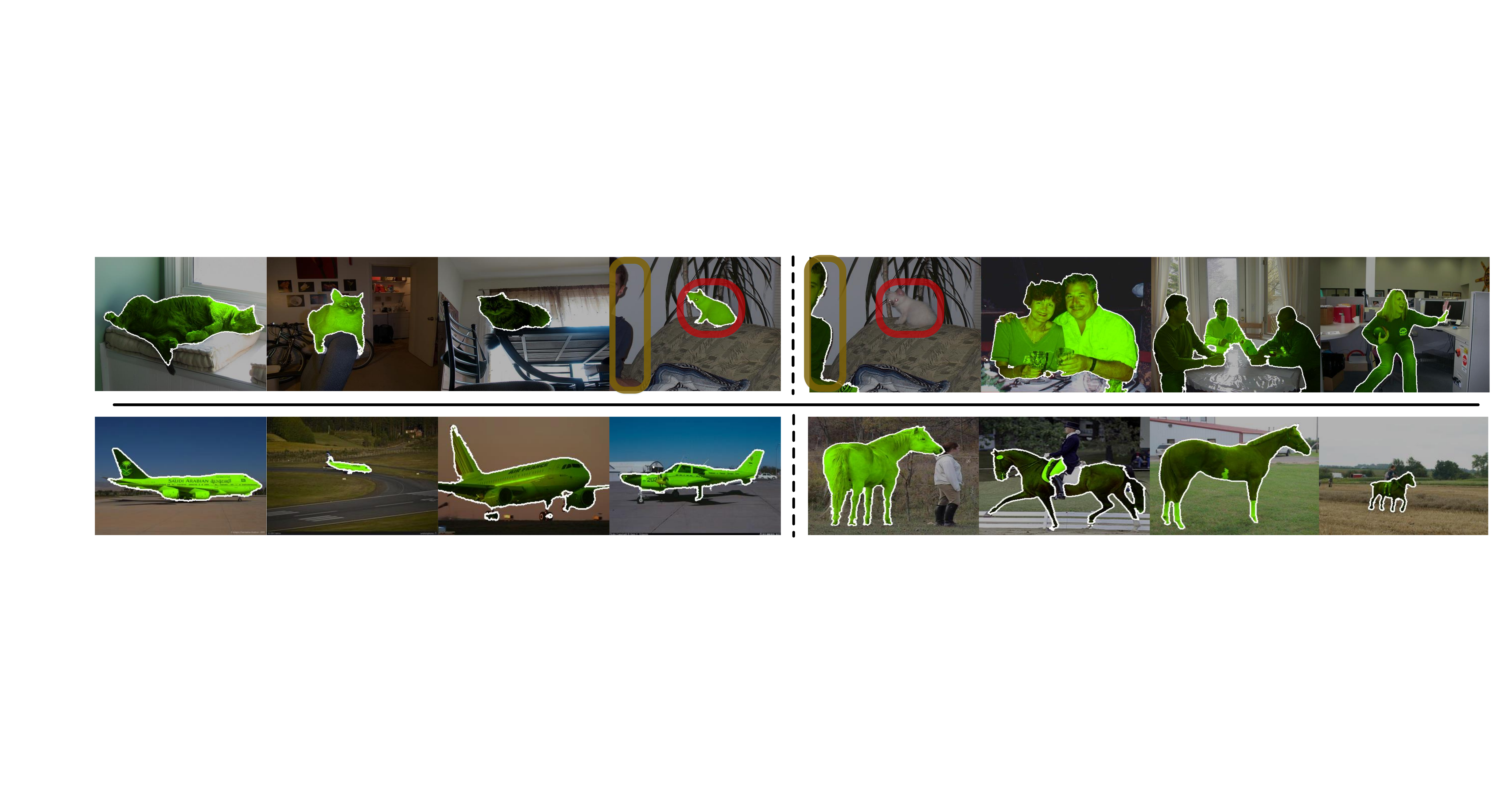}
	\vspace*{-8pt}	
	\caption{\small  Qualitative image object co-segmentation results on PASCAL VOC~\cite{pascal-voc-2012} (top) and Internet~\cite{Rubinstein_2013_CVPR} (bottom). See \S\ref{sec:exvisIOCS}.
	}
	\label{fig:co-seg}	
\vspace*{-14pt}	
\end{figure*}
\vspace*{-6pt}	
\noindent\textbf{Implementation Details:}~Following~\cite{chen2018semantic,DBLP:journals/corr/abs-1804-06423}, we employ  PASCAL VOC to train our model. In each iteration, we randomly sample a group of $N'\!=\!3$ images that belong to the same semantic class, and feed two groups with randomly selected classes (6 images in total) to the network. All other experimental settings are the same as ZVOS.

After training, we evaluate the performance of our method on the test sets of PASCAL VOC and Internet dataset. When processing an image, IOCS must leverage information from the whole image group (as the images are typically different and some are irrelevant)~\cite{quan2016object,vicente2011object}. To this end, for each image $I_i$ to be segmented, we uniformly split the other $N\!-\!1$ images into $T$ groups, where $T\!=\!(N-1)/(N'-1)$. Then we feed the first image group and $I_i$ to a batch of size $N'$, and store the node state for $I_i$. After that, we feed the next group and the store node state of $I_i$ to get a new state of $I_i$. After $T$ steps, the final state of $I_i$ contains its relationships to all other images and is used to produce its final co-segmentation result.

\vspace*{-12pt}
\subsubsection{Quantitative Performance}\label{sec:exqIOCS}
\vspace*{-4pt}
\noindent\textbf{PASCAL VOC.}  It is very challenging to segment the common objects in this dataset, since the objects undergo drastic variation  in scale,  position and appearance. In addition, some images have multiple objects belonging to different categories.  On this dataset, we compare AGNN with six representative methods, including Siamese-based co-segmentation methods~\cite{chen2018semantic,DBLP:journals/corr/abs-1804-06423}, as well as deep semantic segmentation models (\eg., FCNs~\cite{long2015fully}).

Table~\ref{pascal} shows detailed results in terms of mean $\mathcal{J}$. FCNs~\cite{long2015fully} segment each image individually (without considering other related images), and thus give poor performance.  
Both~\cite{chen2018semantic} and~\cite{DBLP:journals/corr/abs-1804-06423} consider pairs of images and gain better results. Our AGNN achieves the best performance because it considers high-order information from multiple images during inference, enabling it to capture richer semantic relations within the image groups.

\noindent\textbf{Internet.}  We evaluate our model (pre-trained on PASCAL VOC) on Internet~\cite{chen2018semantic,quan2016object}.
Quantitative results in Table~\ref{Internet}  again demonstrate the superiority of AGNN (4.5\% performance gain compared with the second best method). The result of AGNN is higher than compared methods for  three  classes: \textit{Car} (84.0\%), \textit{Horse} (72.6\%), \textit{Airplane} (76.1\%). 
\vspace*{-12pt}
\subsubsection{Qualitative Results}\label{sec:exvisIOCS}
\vspace*{-4pt}
Fig.~\ref{fig:co-seg} shows some sample results.  Specifically, the first four images in the top row belong to the \textit{Cat} category (red circle), while the last four images contain the \textit{Person} category (yellow circle) with significant intra-class variation. For both cases, our AGNN successfully detects the common object instances amongst background clutter. For the second row, AGNN also performs well in the cases with remarkable intra-class appearance change.

\begin{table}
	\centering
	\resizebox{0.45\textwidth}{!}{
		\setlength\tabcolsep{3pt}
		\renewcommand\arraystretch{1.0}
		\begin{tabular}{|c|c||cc|}
			\hline\thickhline
			\multirow{2}{*}{Components} &	\multirow{2}{*}{Module} &\multicolumn{2}{c|}{DAVIS$_{16}$}\\
			& &mean $\mathcal{J}$ & $\Delta$$\mathcal{J}$\\
			\hline
			\hline 
			Reference&\textbf{Full model} (3 Iterations, N'= 5)  &80.7 &-\\
			\hline
            \hline
			\multirow{2}{*}{\tabincell{c}{Graph\\Structure}} &\textit{w/o.} AGNN  & 72.2 & -8.5\\
			&\textit{w/o.} Gated Message (Eq.~\ref{aggregation}) & 80.1 & 0.6\\
\hline
			\multirow{3}{*}{\tabincell{c}{Message\\Passing} } & 1 iteration & 78.7 &-2.0\\
			&2 iterations & 79.1 & -1.6\\
			&4 iterations & 80.7 & 0.0\\ \hline
			\multirow{3}{*}{\tabincell{c}{Input\\Frames}} & N'= 3 & 79.6&-1.1\\
			& N'= 6 & 80.7& 0.0\\
			& N'= 7 & 80.7 &0.0\\ \hline		
			Post-Process& 	\textit{w/o.} CRF  & 78.9 & -1.8\\
			\hline \thickhline
	\end{tabular}}	
	\vspace*{-6pt}
	\caption{\small Ablation study (\S\ref{sec:ablation}) on the val set of DAVIS$_{16}$~\cite{perazzi2016benchmark}.}
	\label{table:abl}
	\vspace*{-16pt}	
\end{table}

	\vspace*{-5pt}
\subsection{Ablation Study}\label{sec:ablation}
	\vspace*{-3pt}	
We perform an ablation
study on DAVIS$_{16}$~\cite{perazzi2016benchmark} to investigate the effect of each essential component of AGNN.

\noindent\textbf{Effectiveness of Our AGNN.} To quantify the contribution of our AGNN, we derive a baseline \textit{w/o.~AGNN}, which indicates the results from our backbone model, DeepLabV3. As shown in Table~\ref{table:abl}, AGNN indeed brings significant performance improvements (72.2$\rightarrow$80.6 in term of mean $\mathcal{J}$).

\noindent\textbf{Gated Message Aggregation Strategy.} In Eq.~\ref{aggregation}, we equip the message passing with a channel-wise gated mechanism to decrease the negative influence of irrelevant frames. To evaluate this design, we offer a baseline \textit{w/o.~Gated Message}, which aggregates messages directly. A performance degradation is observed after excluding the gates.

\noindent\textbf{Message Passing Iterations $K$.} To investigate the message passing iterations  $K$, we report the performance as a function of $K$s. We find that, with more iterations ($1\!\rightarrow\!3$), better results can be obtained. The performance of the message passing converges at $K\!=\!3$.

\noindent\textbf{Node Numbers $N'$ During Inference.}
To evaluate the impact of the number of nodes $N'$ during inference, we report performance with different values of $N'$.  We observe that, with more input frames ($3\!\rightarrow\!5$), the performance raises accordingly.  When even more frames are considered ($5\!\rightarrow\!7$), the final performance does not change obviously. This may be due to the redundant content in video sequences.

	\vspace*{-10pt}
\section{Conclusion}
	\vspace*{-4pt}	
This paper proposes a novel AGNN based ZVOS framework for capturing the relations among videos frames and inferring the common foreground objects. It leverages an attention mechanism to capture the similarity between nodes and performs recursive message passing to mine the underlying high-order correlations. 
Meanwhile, we demonstrate the generalizability of AGNN  by extending it to IOCS task. Extensive experiments on three ZVOS and two IOCS datasets indicate that our AGNN performs favorably against current state-of-the-art methods. This further illustrates the importance of  AGNN which can capture diverse relations among similar video frames or semantically related images. 

{\small\noindent\textbf{Acknowledgements} This work was supported in part by ARO grant W911NF-18-1-0296, Beijing Natural Science Foundation under Grant 4182056, CCF-Tencent Open Fund, Zhijiang Lab's International Talent Fund for Young Professionals, and the National Science Foundation (CAREER IIS-1253549).}

{\small
\bibliographystyle{ieee_fullname}
\bibliography{egbib}
}

\end{document}